\documentclass[10pt,twocolumn,letterpaper]{article}

\usepackage{cvpr}              
\usepackage{paralist}
\usepackage{array}
\usepackage{csquotes}
\usepackage{colortbl}
\usepackage{url}
\usepackage[accsupp]{axessibility}

\usepackage[dvipsnames]{xcolor}

\definecolor{cvprblue}{rgb}{0.21,0.49,0.74}
\definecolor{BackgroundColor}{RGB}{247, 217, 250}
\usepackage[pagebackref,breaklinks,colorlinks,citecolor=cvprblue]{hyperref}

\newcommand{\mypartitle}[2][2.5]{\vspace*{-#1 ex}~\\{\noindent {\bf #2}}}
\newcolumntype{P}[1]{>{\centering\arraybackslash}p{#1}}
\newcolumntype{M}[1]{>{\centering\arraybackslash}m{#1}}
\newcolumntype{L}[1]{>{\arraybackslash}m{#1}}
\newcommand\blfootnote[1]{%
  \begingroup
  \renewcommand\thefootnote{}\footnote{#1}%
  \addtocounter{footnote}{-1}%
  \endgroup
}

\def\paperID{5} 
\def\confName{CVPR}
\def\confYear{2024}

\title{Exploring the Zero-Shot Capabilities of Vision-Language Models\\ for Improving Gaze Following}

\author{Anshul Gupta* \and
\and
Pierre Vuillecard*
\and
Arya Farkhondeh
\and
Jean-Marc Odobez
\and
Idiap Research Institute, Martigny, Switzerland\\
École Polytechnique Fédérale de Lausanne, Switzerland\\
{\tt\small \{agupta, pvuillecard, afarkhondeh, odobez\}@idiap.ch}
}

\begin{document}
\maketitle
\blfootnote{* indicates equal contribution}
\begin{abstract}
Contextual cues related to a person's pose and interactions with objects and other people in the scene can provide valuable information for gaze following. While existing methods have focused on dedicated cue extraction methods, in this work we investigate the zero-shot capabilities of Vision-Language Models (VLMs) for extracting a wide array of contextual cues to improve gaze following performance. We first evaluate various VLMs, prompting strategies, and in-context learning (ICL) techniques for zero-shot cue recognition performance. We then use these insights to extract contextual cues for gaze following, and investigate their impact when incorporated into a state of the art model for the task. Our analysis indicates that BLIP-2 is the overall top performing VLM and that ICL can improve performance. We also observe that VLMs are sensitive to the choice of the text prompt although ensembling over multiple text prompts can provide more robust performance. Additionally, we discover that using the entire image along with an ellipse drawn around the target person is the most effective strategy for visual prompting. For gaze following, incorporating the extracted cues results in better generalization performance, especially when considering a larger set of cues, highlighting the potential of this approach.
\end{abstract}    
\section{Introduction}
\label{sec:intro}

\begin{figure}[t]
\includegraphics[width=1.0\linewidth]{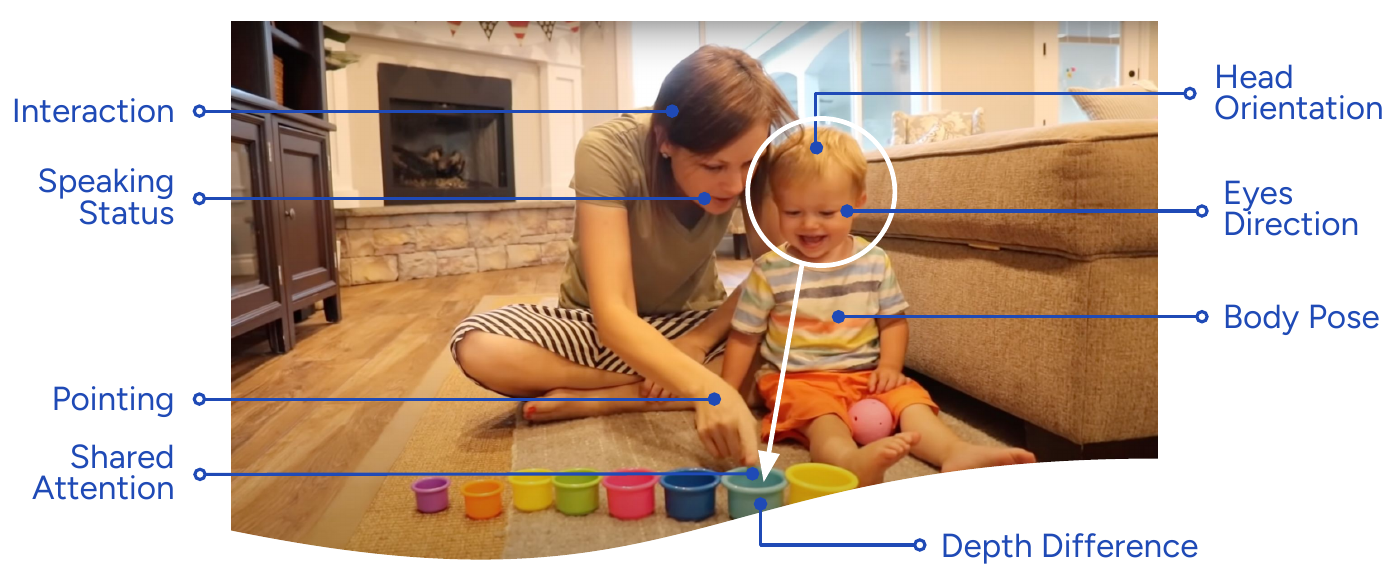}
\caption{ As humans, we rely on various sources of information to predict a person's gaze target. This image shows contextual information that could be valuable.}
\label{fig:context_gaze}
\end{figure}

Understanding where a person is looking in a scene, also known as gaze following, has diverse applications, including human-robot interaction~\cite{sheikhi2015combining, jin2022depth, admoni2017social}, conversation analysis~\cite{otsuka2011conversation, gatica2009automatic_conversation}, and the study of neurodevelopmental disorders~\cite{chong2020dvisualtargetattention, li2022appearance}. However, this is a challenging task, demanding a model to interpret a large spectrum of contextual cues such as the person's interactions with objects and other people in the scene. 
For instance, it has been shown that eye and hand movements are coordinated during manipulation activities~\cite{Johansson2001}. Or that during conversations, people usually look at the person talking~\cite{Stiefelhagen1999} and that leveraging this information can help in gaze target selection in meeting settings~\cite{Otsuka:ICMI:2005}. As seen in Figure~\ref{fig:context_gaze}, estimating the child's gaze target requires understanding their head and body pose, the interaction between the child and the adult through pointing and shared attention etc.

In order for a model to capture these cues and learn their impact on gaze target selection, it would need to be trained on a high-quality and large-scale labeled dataset. However, existing gaze following datasets~\cite{chong2020dvisualtargetattention, tafasca2023childplay} are small-scale, hindering the effective utilization of these cues. To address these challenges, prior works have relied on dedicated cue extraction methods such as inferred body pose~\cite{gupta2022modular, lian2018believe}, and supplied them to models for improved performance. However, these approaches focus on specific cues and do not supply the larger array of contextual cues which could be needed for accurate gaze target prediction. Traditional methods to address this limitation include: (1) manually annotating for relevant cues; but it is cost-intensive and not always available during inference, or (2) pseudo-labelling with expert models; however this requires access to a multitude of task specific models for each cue or a subset of cues. Hence, it is evident that novel solutions are required. 

Given these challenges, we investigate the potential of Visual Language Models (VLMs) to extract valuable contextual cues for gaze following, aiming to overcome the constraints of traditional labeling approaches. VLMs have shown promising zero-shot performance for a variety of tasks~\cite{pmlr-v139-radford21a, zhang2023vision}, owing to their ability to learn visual-text associations at scale. Hence, a single model may be capable of extracting all relevant contextual cues. At the same time, given the zero-shot setting, the set of cues to be considered can be adjusted based on the domain, further increasing this approach's applicability. 

In this work, we consider cues related to pose, person-person interactions, and person-object interactions. We first evaluate the zero-shot performance of different VLMs for recognising these cues (Section~\ref{sec:contextual_cues}), and leverage the best performing approach to extract them. We then investigate whether incorporating these extracted cues can improve gaze following performance (Section~\ref{sec:text_improved_gaze}).



\mypartitle{Challenges.}
While VLMs have shown impressive zero-shot performance for a variety of tasks, these tasks (ex. image classification) usually involve processing the entire image. However, for accurate gaze following we also need to capture contextual cues related to each person in the scene. Hence, we need to consider an appropriate visual prompt to allow the VLM to focus on the person of interest. At the same time, it is important to consider the choice of text prompt as VLMs have been shown to benefit from prompt engineering~\cite{radford2018improving}. Finally, given the extracted cues from the VLMs, we need to consider how to incorporate such information into a gaze following model. 
%
Following these research questions, we make the following contributions:
\begin{compactitem}
    \item \textit{VLMs for contextual cues extraction}: We explore 4 state of the art VLMs~\cite{pmlr-v139-radford21a, li2022blip, li2023blip} for this task. We also investigate different visual prompts to focus on the person of interest, and different text prompts to describe the cue of interest. We show that VLMs can indeed capture contextual cues although the choice of VLM, visual prompt and text prompt impacts performance.
    \item \textit{Text improved Gaze Following}: We incorporate the extracted contextual cues into a recent transformer based gaze following model~\cite{gupta2023_gazeinteract}. Our results indicate that incorporating these cues can result in better generalization performance, especially when considering larger sets of cues.
\end{compactitem}

\section{Related-Work}
\label{sec:rel_work}

\mypartitle{Vision-Language Models (VLMs).} VLMs began receiving significant attention following the introduction of CLIP (Contrastive Language-Image Pretraining)~\cite{pmlr-v139-radford21a}. This contrastive learning framework learns effective multi-modal representations using image-text pairs. CLIP demonstrated impressive zero-shot classification performance on standard image classification benchmarks. Subsequent VLMs, such as BLIP~\cite{li2022blip} and BLIP-2~\cite{li2023blip}, have been introduced with notable differences from CLIP. These include varied objectives (incorporating additional losses beyond the contrastive approach), the use of more curated training datasets, and enhanced image caption generation through caption filtering. Specifically, BLIP-2 has introduced advanced pre-training strategies, integrating a frozen image encoder and a Querying Transformer (Q-Former), which enables a nuanced extraction of visual representations. They have even shown strong performance for video tasks such as action recognition~\cite{pmlr-v139-radford21a} and text-to-video retrieval~\cite{li2022blip} despite not processing the temporal dimension. However, their performance for localizing the actions/cues of people in an image has not been explored.

While there has been some work on video language models, they face certain limitations. Firstly, available video-text pairs for pretaining is limited compared to the scale of available image-text pairs \cite{xu2021videoclip}, so these methods do not generalize as well. To cope with this issue, some works leveraged pre-trained image-based VLMs and adapted them for video input, however, it harmed their zero-shot performance \cite{wang2021actionclip, ju2022prompting}. Secondly, these models are computationally more expensive, and cannot be applied on static gaze following datasets.

More recently, VLMs such as BLIP2 have leveraged Large language models (LLMs) for generating textual output. LLMs have shown an impressive ability to act as models of the world, with a rudimentary understanding of agents, beliefs and actions~\cite{andreas2022language_world}, and an ability to perform commonsense and mathematical reasoning~\cite{chowdhery2023palm_world}. Hence, they may also be capable of capturing complex relationships between people and objects in the scene to better extract contextual cues for human behaviour understanding. Further, they have been shown to benefit from in-context learning~\cite{brown2020language_gpt3} (ICL), wherein a few demonstrations of the task are provided at inference time without any weight updates. VLMs that exploit such LLMs have also displayed improvements in performance using ICL~\cite{alayrac2022flamingo, tsimpoukelli2021multimodal_icl}, by being provided with a few sample visual and/or textual demonstrations. It is still a research question whether LLMs and ICL can improve the recognition of gaze contextual cues.

\mypartitle{Visual and Textual Prompting.}
The recent work of Shtedritski et al.~\cite{Shtedritski_2023_ICCV} explored different visual prompting approaches for CLIP. They compared cropping the visual area of interest versus drawing a red ellipse around it, and found the red ellipse approach to perform better for keypoint localization and referring expression comprehension tasks. They also observed that blurring or graying the region outside the ellipse can provide additional benefits. However, the performance of these approaches for action/cue recognition hasn't been explored. On the textual prompting side, the original CLIP paper~\cite{pmlr-v139-radford21a} showed that prompt engineering, such as using the prompt 'a photo of a \{label\}' improved performance over using just the label text on ImageNet~\cite{russakovsky2015imagenet}. They also observed that ensembling different prompts in the embedding space can reliably improve performance.
Recent works~\cite{zhou2022learning, zhou2022conditional} introduced learnable prompts that were fine-tuned on a specific task. However this approach requires data to adapt, and hence cannot be applied in a zero-shot manner. Also, although the learnable prompts can then be included with prompts for unseen classes, the results lag behind manual prompt engineering efforts \cite{zhou2022conditional}. In this work, we evaluate different manual prompt engineering approaches that can be applied to any new set of classes.

\mypartitle{Contextual Cues for Gaze Following.}
Previous works in estimating the Visual Focus of Attention (VFOA) of a person leveraged cues such as the head pose and speaking information of people~\cite{Otsuka:ICMI:2005, sheikhi2015combining, StiefelhagenFocusMultiCues2002} for improved performance. However, these methods typically require access to frontal views of people and knowledge of the 3D scene structure (ex. using multiple calibrated cameras) for inference. This makes it challenging to deploy these algorithms in new environments where such information may not be available.

Hence, Recasens et al.~\cite{recasens2017following} proposed the Gaze Following task to estimate the scene gaze target of a person in an image using only the image and with no prior assumptions about the scene or camera placement. However, due to the complexities of scenes and the lack of data, models have encountered challenges in capturing pertinent information, ultimately leading to sub-optimal predictions. Hence, recent methods have shown that inferred cues such as depth~\cite{Fang_2021_CVPR_DAM, tafasca2023childplay, gupta2022modular, jin2022depth, hu2022we, bao2022escnet, tonini2022multimodal} and body pose~\cite{lian2018believe, gupta2022modular} can be leveraged for improved performance. However, incorporating person-specific auxiliary information in these models is not straightforward.

More recently, \cite{gupta2023_gazeinteract} proposed a new transfomer based model for gaze following and social gaze prediction. They showed that this architecture allows for easily incorporating person-specific auxiliary cues, with improved performance from the addition of people's speaking status. Whether it can benefit from additional cues remains to be explored.
\section{Contextual Cues Extraction}
\label{sec:contextual_cues}

In the first stage, we evaluate the zero-shot performance of different VLMs and prompting approaches for recognition of cues. Note that we interchangeably refer to a specific cue as a class.

\subsection{Method}
\label{sec:contextual_cues_method}

We investigate different visual and textual prompting strategies, as well as two different variants of VLMs for zero-shot contextual cues extraction, namely image-text matching (ITM) and visual question-answering (VQA). 

\mypartitle{Visual Prompting.} In a complex scene involving multiple people,  ITM becomes challenging as our task requires conditioning on a specific target person. To address this, we investigate various visual prompting techniques that enable ITM to focus on a chosen individual. We employ several approaches, including no prompting, drawing a red ellipse around the person following~\cite{shtedritski2023does_red_circle}, blurring or graying the background. These techniques are applied either to the entire image (image-based) or to the cropped target person (person-based), resulting in a total of eight distinct visual prompting approaches. We provide an example of the different visual prompts in Figure~\ref{fig:visual_prompt} in the supplementary.

\mypartitle{Text Prompting.} In our approach to text prompting, we employed a structured method for generating prompts systematically based on templates. This method allows us to meticulously examine the impact of each textual component within the prompt. A template, in this context, is a fixed sentence where only specific parts can be altered. For examples, "a \{photo\} of a \{person\} \{class\}", and "a \{person\} is \{class\}", are two instances of templates. Beyond the varied sentence structures, the placeholders \{photo\}, \{person\}, and \{class\} can be substituted with semantically related components. For instance, \{photo\} could be replaced with "picture" or "snapshot," \{person\} might be substituted with "individual" or "human," and \{class\} can refer to class synonyms such as "talking" or "narrating" if the original class is "speaking". In this work, synonyms refer to changes in class synonyms otherwise mentioned explicitly. In the supplementary, Figure~\ref{fig:text_prompt} presents all the different templates and synonyms used in this section. 

\mypartitle{Image-Text Matching (ITM).}
In ITM, the objective is to compute a cosine similarity between the visual and textual embedding. A high similarity suggests that the image contains the textual description. Formally, given an image $I$ of size $H \times W \times 3$ and a set of K class names, we use the visual and text encoders of a pre-trained VLM (e.g., CLIP) to get a visual embedding $e_{I} \in \mathbb{R}^{d}$ and K text embeddings $e_{T} \in \mathbb{R}^{K \times d}$. We perform the following matching:

\begin{equation}
    S = dot (e_{I} \cdot e_{T}^{\mathrm{T}}) 
\end{equation}

\noindent where, $S \in \mathbb{R}^K$ are the resulting similarity scores. When multiple textual prompts refer to the same class name k, i.e. $e_{T_k} \in \mathbb{R}^{P \times d} $, we can perform an \textit{Ensemble} to get the score.

\begin{equation}
    S_k = dot (e_{I} \cdot \frac{1}     {P}\sum_{e\in e_{T_k}}e^{\mathrm{T}}) 
\end{equation}

\noindent The \textit{Ensemble} approach utilizes the mean embedding, acting as a centroid for a given class and thus is expected to be more robust.
The scores for each class are then normalized across samples to have a zero mean and a standard deviation of one.
In this work, we investigate three different pre-trained VLMs such as CLIP~\cite{pmlr-v139-radford21a}, BLIP~\cite{li2022blip} and BLIP-2~\cite{li2023blip}. For more details regarding these models, we refer the readers to the original papers and details in Sec.~\ref{sec:rel_work}.

\mypartitle{Visual Question Answering (VQA).}
In order to explore the potential of LLMs for our task, we investigate a recent VQA model, BLIP-2 VQA~\cite{li2023blip}, that leverages a LLM called FlanT5~\cite{raffel2020exploring_t5}. In VQA models, a textual question is jointly input with an image to the model, and the model outputs a textual answer. We convert the text prompts described previously into a set of questions that result in simple \enquote{yes} or \enquote{no} answers, which we then convert into a binary score. Examples of prompts are displayed in supplementary Fig~\ref{fig:vqa_prompt}. To further explore the benefits of ICL, we provide additional textual context in the form of a generated caption from the same model. Thus, the text input to the model is of the form \textit{\{generated caption\} \{text prompt\}}. 
It is worth noting that the BLIP-2 VQA model is much slower to run than the ITM models as (1) the model is much larger due to the LLM, and (2) the answer is conditioned on the image \textit{and} question, so we need to run a forward pass for each image-prompt pair. This is unlike the ITM models where the images and prompts can be processed separately, with a similarity score computed afterward.

\subsection{Experiments}

\mypartitle{Datasets.} 
We employ two datasets to shed light on the VLMs' ability to extract meaningful cues. 

\textit{ChildPlay:} We manually annotated 6 cues from the ChildPlay~\cite{tafasca2023childplay} dataset, which is a recently proposed dataset for gaze following. For each class, we selected around 50 clear positives and 50 clear negatives. The classes and statistics are presented in the supplementary Table~\ref{tab:childpay_stat}.

\textit{AVA-Actions:} Then, to scale our evaluation we used the validation split of the AVA dataset~\cite{Gu_2018_CVPR}, which is a human action localization dataset. This dataset is much more challenging since it is heavily unbalanced and large scale containing around 41000 images. A subset of the classes of interest was selected. In Table \ref{tab:ava_results}, a summary of the dataset classes and distribution is shown.

\begin{table}[t]
\resizebox{\linewidth}{!}
{
    \centering
    \begin{tabular}{m{8cm} c}
        \toprule
        \textbf{Selected Classes - AVA} & \textbf{Support} \\
        \midrule
        \textbf{Pose (P)} \\ 
        stand & 23424 \\
        sit & 16660\\
        bend/bow (at the waist) & 1512 \\
        \midrule
        \textbf{Person-Person Interaction (P-P)} \\ 
        talk to (e.g., self, a person, a group) & 25985 \\
        hug (a person) & 340 \\
        hand clap & 330 \\
        give/serve (an object) to (a person) & 313 \\
        \midrule
        \textbf{Person-Object Interaction (P-O)} \\ 
        carry/hold (an object) & 17199\\
        touch (an object) & 5099\\
        read & 658\\ 
        write & 273\\
        lift/pick up & 118\\
        text on/look at a cellphone & 112 \\
        work on a computer & 111\\
        \bottomrule
    \end{tabular}
}
\caption{Selected classes from the AVA Dataset (validation set) categorized as Pose (P), Person-Person Interaction (P-P), and Person-Object Interaction (P-O), including the number of samples (support) for each.} 
\label{tab:ava_results}
\end{table}

\mypartitle{Metrics.}
We leverage two metrics:
\begin{compactitem}
    \item \textit{AP}: To evaluate the performance of different VLMs and prompting approaches, we use Average Precision (AP). It is computed per class between the ground truth and the scores obtained from the VLMs. We also consider the mean of the AP scores across all classes or mean Average Precision (mAP).
    \item \textit{Accuracy}: Since the output of the VQA variants is a binary decision, we cannot compute AP; instead, we compute accuracy. To compare with ITMs, we binarize their output by applying a threshold of zero since the scores are normalized with a zero mean (however they may benefit from optimizing the threshold).
\end{compactitem}

\subsection{ITM Results}

\mypartitle{Visual prompting.}
\label{sec:visual prompting}
\begin{figure}[t]
    \centering
    \includegraphics[width=1.0\linewidth]{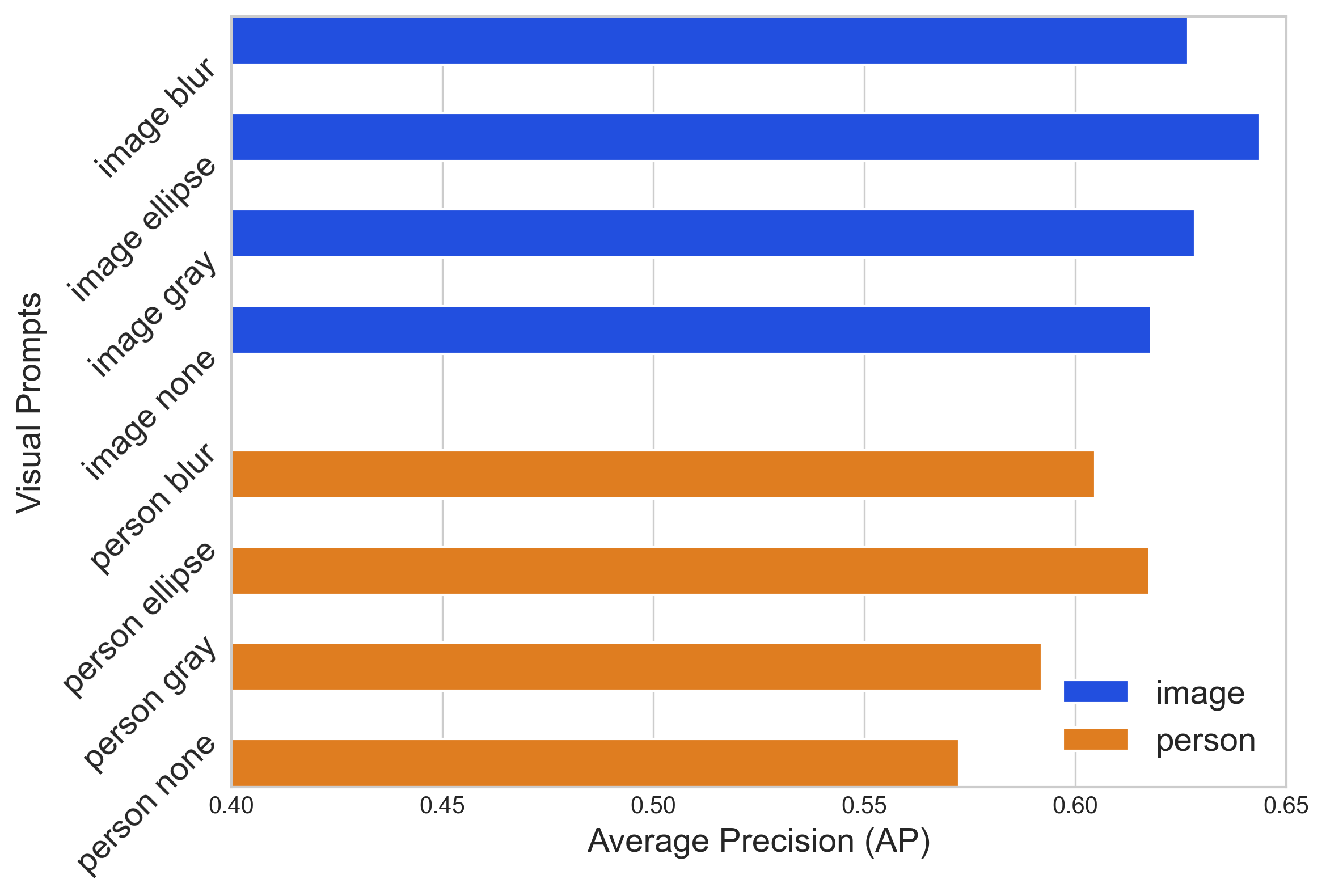} 
    \caption{Results of different visual prompting approach on Childplay. \textit{Image} corresponds to input the full image whereas \textit{person} refers to the use of person crop as input.}
    \label{fig:results_visual}
    \vspace{-2mm}
\end{figure}
We compare the performance of the different visual prompts described in Section~\ref{sec:contextual_cues_method} in Fig.~\ref{fig:results_visual}. The results are aggregated across VLMs and different text prompts, and categorized by the type of visual input, i.e. image-based versus person-based. We see that image-based approaches outperform person-based variants. This suggests that a broader visual input provides additional context, enhancing the zero-shot recognition for the target person in the image. Furthermore, among the visual prompts, the red ellipse approach outperforms others, aligned with findings in \cite{shtedritski2023does_red_circle}. Therefore, in subsequent experiments, we employ the image-based red ellipse as the visual prompt.


\mypartitle{VLMs.}
\begin{figure}[t]
    \centering
    \includegraphics[width=1.0\linewidth]{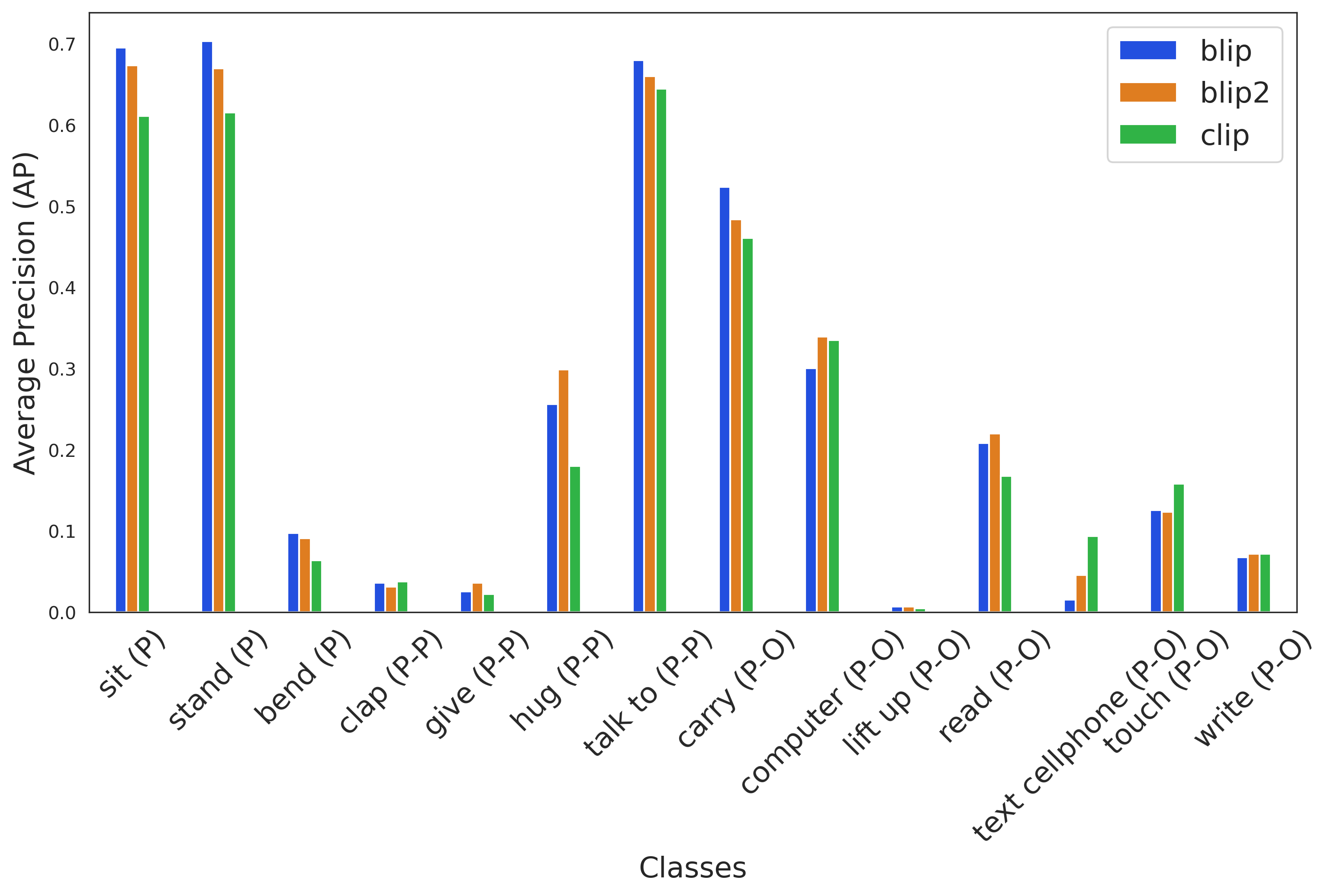}
    \caption{Results of different VLMs following the ITM approach on AVA. Three VLMs are compared across different classes categorized as Pose (P), Person-Person Interaction (P-P), and Person-Object Interaction (P-O).}
    \label{fig:results_vlms}
\end{figure}
We compare the performance of three VLMs, namely CLIP, BLIP, and BLIP-2. In Fig.~\ref{fig:results_vlms}, we present a class-wise comparison of the three VLMs on AVA. Note that, for each VLM, we aggregate the results from different textual prompts. Firstly, we observe that no single model always outperforms the others. However, BLIP and BLIP-2 surpass CLIP in pose and person-to-person classes, while CLIP performs well when the class refers to a clear object, such as \textit{work on a computer} or \textit{text on a cellphone}. This may be related to differences in training data, and is a direction for investigation. On average, BLIP-2 is the top performing model. In the subsequent analysis, we continue focusing on BLIP-2 while varying the text prompting aspects. 


\mypartitle{Text prompting.}
\label{sec:text prompting}
\begin{figure}[t]
    \centering
    \includegraphics[width=1.0\linewidth]{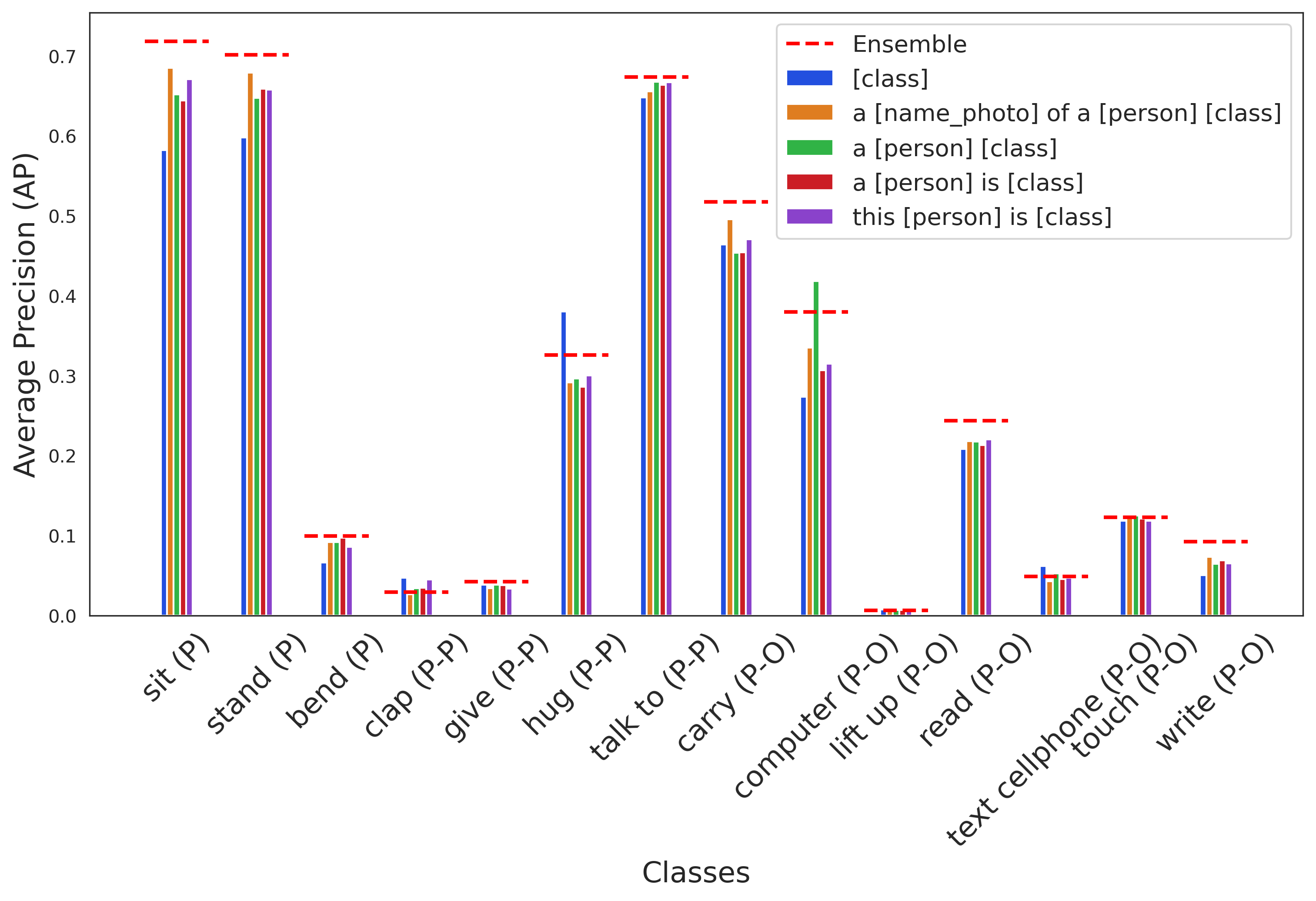}
    \caption{Results of different templates using BLIP-2 on AVA. Six templates are compared across different classes. } 
    \label{fig:results_templates}
    \vspace{-2mm}
\end{figure}
%
We investigate the impact of the text prompts described in Section~\ref{sec:contextual_cues_method} at two different levels, at the template level and synonym level. When evaluating the template, we aggregate results over the other text prompt component variations. Similarly for when we evaluate the class synonym.
In Figure \ref{fig:results_templates}, performance for different templates are shown on AVA per class. Firstly, there is no best template overall, which correlates to the finding of~\cite{pmlr-v139-radford21a} that VLMs are prompt sensitive. However, using the \textit{Ensemble} approach described in Section~\ref{sec:contextual_cues_method} provides more robust performance, often outperforming the best template, and always outperforming the worst template. In addition, the wording in textual prompts matters, as can be seen in the supplementary Figure~\ref{fig:synonym}, where different class synonyms can change the performance by a large margin. However, we notice that for most of the classes, including \textit{\{person\}} in the prompt improves performance. This suggests that conditioning the prompt to an individual helps to extract person-centric information.

\subsection{VQA Results}

\begin{figure}[t]
    \centering
    \includegraphics[width=1.0\linewidth]{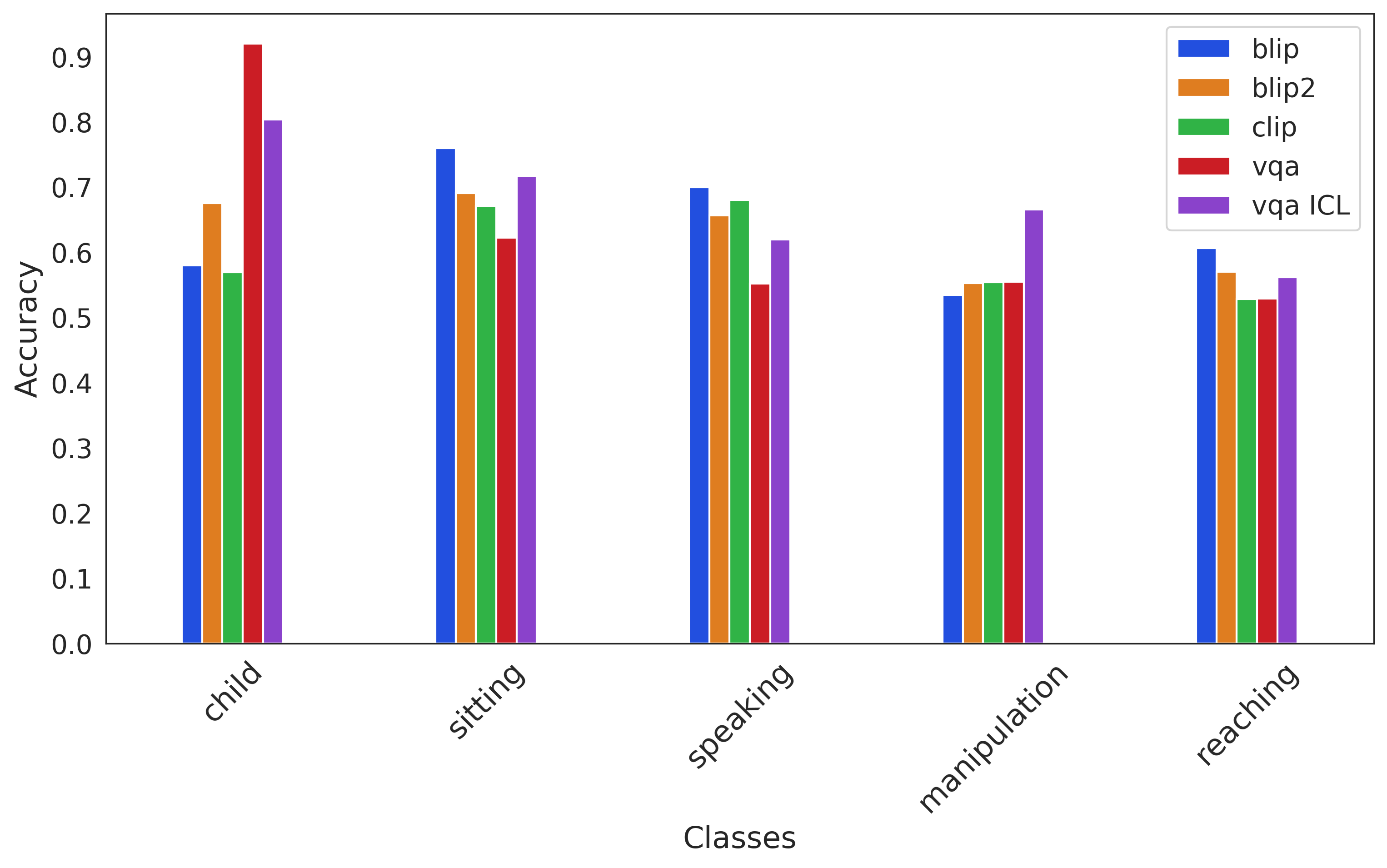}
    \caption{Results of BLIP-2 vqa with and without in-context learning, vqa ICL and vqa respectively, on ChildPlay. It is compared with the VLMs CLIP, BLIP and BLIP-2.}
    \label{fig:results_vqa}
    \vspace{-2mm}
\end{figure}

To investigate the potential of LLMs and in-context learning for contextual cues extraction, we evaluate the BLIP-2 VQA model on the ChildPlay dataset, and compare it against ITM based VLM models (Figure~\ref{fig:results_vqa}). Note that the results are aggregated across all text prompts. As mentioned in Section~\ref{sec:contextual_cues_method}, the BLIP-2 VQA model is much slower to run compared to the ITM based models which is why we use the smaller ChildPlay dataset. We also use a smaller set of templates and synonyms in the text prompt (Fig.~\ref{fig:vqa_prompt} in supplementary) to reduce computation time. 

\mypartitle{Benefit of LLM.}
Comparing the performance of BLIP-2 against BLIP-2 VQA (BLIP-2 and vqa in the figure), we see that BLIP-2 VQA does much better for the 'child' class, but on par of worse for the other classes. This suggests that the LLM in the BLIP-2 VQA model is not necessarily providing better results. However, as mentioned previously, this model uses a smaller set of templates and synonyms in the text prompt for computational reasons so may benefit from using a larger set.

\mypartitle{In-Context Learning.}
We see that the BLIP-2 vqa model with ICL improves for all classes except the 'child' class compared to no ICL. This is in contrast to the observations in the original paper where the architecture is introduced~\cite{li2023blip}, and suggests the potential of leveraging ICL for contextual cues extraction.

\section{Text-Improved Gaze Following}
\label{sec:text_improved_gaze}

In the second stage, we apply insights from Section~\ref{sec:contextual_cues} and leverage BLIP-2 along with the red ellipse visual prompting approach, and the \textit{Ensemble} text prompting approach to extract contextual cues. We then evaluate the impact of incorporating these cues into a gaze following model. 

\subsection{Method}
\label{sec:text_improved_gaze_method}

We employ the static version of the recently proposed MTGS~\cite{gupta2023_gazeinteract} model. This model is a transformer-based architecture designed for multi-person gaze following and social gaze prediction. Given an input image and head crops of people in the scene, it first produces two types of tokens: image tokens ($\text{x}_{\text{image}} \in \mathbb{R}^{N \times D}$), similar to those in a standard Vision Transformer (ViT) architecture~\cite{dosovitskiy2020image}, and person gaze tokens ($\text{x}_{\text{gaze}} \in \mathbb{R}^{P \times D}$), where $P$ represents the number of people in the scene. Person tokens are generated using head crops, a gaze backbone, and a subsequent linear projection layer. This formulation naturally supports incorporating contextual cues for each person, as the information can be fused with the corresponding person token.

\begin{figure}[t]
    \centering
    \includegraphics[width=1.0\linewidth]{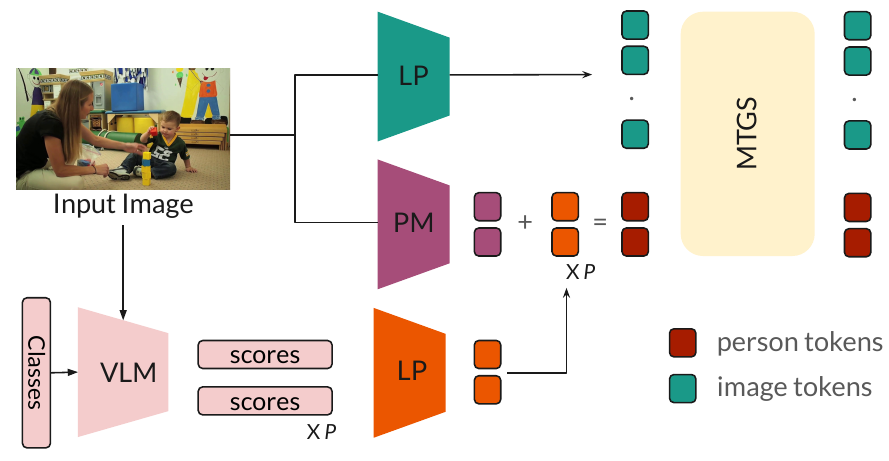}
    \caption{An overview of Text-Improved Gaze Following: Given an image containing $P$ persons, image tokens and person tokens are generated via a Linear Projection (LP) and a person module (PM) respectively. To incorporate VLM contextual information, we use a VLM to obtain $P$ score vectors, each with the dimension as the number of classes ($K$). We then linearly project these vectors and perform early fusion by adding them to the corresponding person tokens. Scene and updated person tokens are subsequently passed to MTGS~\cite{gupta2023_gazeinteract} to model person and scene interactions using self and cross-attention modules across multiple blocks.}
    \label{fig:text_improved_gf}
    \vspace{-2mm}
\end{figure}

Given the success of additive fusion in the case of position embeddings for transformers~\cite{dosovitskiy2020image}, and early fusion of body pose and depth information for gaze following models~\cite{gupta2022modular}, we aim to incorporate contextual information derived from VLMs in an early fusion and additive manner. To this end, as illustrated in Fig.~\ref{fig:text_improved_gf}, we use a linear projection layer ($\Phi$) to project the vector of predicted scores ($S_{vlm} \in \mathbb{R}^{P \times K}$, $K$ is the number of classes) and generate person context tokens matching the dimensions of the person gaze tokens ($\Phi (S_{vlm}) \in \mathbb{R}^{P \times D}$). We then apply the $add$ operation to combine the person context tokens from the VLMs with the corresponding person gaze tokens. Following this, the enriched person gaze tokens, now with added contextual cues, and the image tokens are fed into MTGS, where, people and scene tokens interact through self and cross-attention modules across multiple blocks.
\begin{equation}
    \text{x}_{\text{out}} = \text{MTGS}\left( \left[ \text{x}_{\text{gaze}} + \Phi \left( \text{S}_{\text{vlm}} \right), \text{x}_{\text{image}} \right] \right)
\end{equation}
Finally, a prediction module takes the updated tokens ($\text{x}_{\text{out}}$) and predicts the visual attention heatmap for each person, as well as pair-wise social gaze labels. For a more comprehensive understanding of the architecture, we direct readers to the original paper~\cite{gupta2023_gazeinteract}.

\subsection{Experiments}
\label{sec:experiments-gf}

\mypartitle{Datasets.}
We leverage two gaze following datasets:

 \textit{GazeFollow}~\cite{recasens2017following} is a large-scale static dataset for gaze following, featuring 122K images. Most images are annotated for a single person with their head bounding box and gaze target point. The test set contains annotations by multiple annotators. Despite lower quality images and annotations, it's diversity makes it a good dataset for pre-training.
 
 \textit{ChildPlay}~\cite{tafasca2023childplay} is a recent video dataset for gaze following, featuring children playing and interacting with other children and adults. It is annotated with the head bounding box, gaze point and gaze label (consisting of 7 non-overlapping classes) of people in the scene.

 Following~\cite{gupta2023_gazeinteract}, we pre-process both datasets to extract pair-wise social gaze labels for two tasks:
 \begin{compactitem}
     \item \textit{LAH:} It stands for looking at humans and occurs when a person looks at another person's head. In terms of positive/negative pair statistics, it is 27k/493k for GazeFollow, and 59k/682k for ChildPlay.
     \item \textit{LAEO:} It stands for looking at each other and occurs when two people engage in mutual gaze. In terms of positive/negative pair statistics, it is 0/0 for GazeFollow, and 7k/351k for ChildPlay.
 \end{compactitem}
 
\mypartitle{Contextual Cues.}
We define three sets of contextual cues:
\begin{compactitem}
    \item \textit{AVA+CP:} These are the set of 24 cues defined in Section~\ref{sec:contextual_cues} for AVA and ChildPlay that were used for evaluating different VLMs and prompting strategies. 
    \item \textit{HICO:} The HICO dataset~\cite{chao2015_hico} is human-object interaction dataset that defines a list of 117 interaction verbs. We leverage these verbs as contextual cues.
    \item \textit{SWIG:} The SWIG-HOI dataset~\cite{Wang_2021_ICCV_swig} is a large-scale human-object interaction dataset that defines 406 verbs. We leverage these verbs as contextual cues.
\end{compactitem}
We provide the manually curated synonyms and templates used for generating different text prompts for AVA+CP in Figures~\ref{fig:text_prompt},\ref{fig:vqa_prompt} of the supplementary. For HICO and SWIG, we use the same set of templates, but generate 4 synonyms for each cue using ChatGPT~\cite{chatgpt}.

\mypartitle{Training and Validation.}
Following~\cite{gupta2023_gazeinteract}, we train the model for 20 epochs on GazeFollow using a learning rate of 1e-4 and the AdamW~\cite{loshchilov2018decoupled_adamw} optimizer. We supervise using the standard MSE loss for gaze heatmap prediction, and binary cross entropy loss for LAH prediction. For validation, we use the split proposed by~\cite{tafasca2023childplay}.

\mypartitle{Metrics.}
We use the standard gaze following metrics:
\begin{compactitem}
    \item \textit{AUC}: the predicted heatmap is compared against a binary GT map with value 1 at annotated gaze point positions, to compute the area under the ROC curve. 
    \item \textit{Distance (Dist.)}: the arg max of the heatmap provides the gaze point. 
    We can then compute the L2 distance between the predicted and GT gaze point on a $1\times 1$ square. We compute Minimum (Min.) and Average (Avg.) distance against all annotations.
\end{compactitem}
In addition, we compute F1 scores for LAH (\textit{F1$_{\text{LAH}}$}) and LAEO (\textit{F1$_{\text{LAEO}}$}). For LAH, we check if the predicted gaze point falls inside the target person's head bounding box. For LAEO, we check the reverse as well. \footnote{The predicted LAH scores can also be used for these tasks but were shown to have slightly lower performance~\cite{gupta2023_gazeinteract}.}

\subsection{Results}
\label{sec:gf-results}

\begin{table}[t]
\resizebox{\linewidth}{!}{
\centering
\begin{tabular}{l | cccc}
\toprule
\textbf{Model} & \textbf{AUC$\uparrow$} & \textbf{Avg. Dist$\downarrow$}  &  \textbf{Min. Dist$\downarrow$} & \textbf{F1$_{\text{LAH}}$$\uparrow$}\\
\midrule
Fang~\cite{Fang_2021_CVPR_DAM} & 0.922 & 0.124 & 0.067 & -\\
Tonini~\cite{tonini2022multimodal} & 0.927 & 0.141 & - & -\\
Jin~\cite{jin2022depth} & 0.920 & \underline{0.118} & 0.063  & -\\
Bao~\cite{bao2022escnet} & 0.928 & 0.122 & -  & -\\
Hu~\cite{hu2022gaze_object} & 0.923 & 0.128 & 0.069  & -\\
Tafasca~\cite{tafasca2023childplay} & 0.936 & 0.125 & 0.064 & -\\
Chong~\cite{chong2020dvisualtargetattention} & 0.921 & 0.137 & 0.077 & -\\
Gupta~\cite{gupta2022modular} & 0.933 & 0.134 & 0.071 & -\\
Jin~\cite{jin2021multi} & 0.919 & 0.126 & 0.076 & -\\
\midrule
MTGS~\cite{gupta2023_gazeinteract}  &  0.929 & \underline{0.118} & 0.062 & \underline{0.639}\\
\rowcolor{BackgroundColor} MTGS + AVA + CP & 0.936 & \underline{0.118} & \underline{0.061} & \textbf{0.643}\\
\rowcolor{BackgroundColor} MTGS + HICO & 0.934 & \textbf{0.116} & \textbf{0.060} & \underline{0.639}\\
\rowcolor{BackgroundColor} MTGS + SWIG & 0.933 & 0.119 & \underline{0.061} & 0.619\\
\bottomrule
\end{tabular}
}
\caption{Results for incorporating VLM context with different sets of classes on the GazeFollow dataset. AVA+CP has 24 classes, HICO has 117 classes and SWIG has 406 classes. Best results are given in bold, second best results are underlined.}
\label{tab:results-gf}
\vspace{-2mm}
\end{table}

\begin{table}[t]
\resizebox{\linewidth}{!}{
    \centering
    \begin{tabular}{L{3cm} | M{1.5cm} M{1.5cm} M{1.5cm}}
        \toprule
       \textbf{Method} & \textbf{Dist.$\downarrow$} & \textbf{F1$_{\text{LAH}}\uparrow$}  & \textbf{F1$_{\text{LAEO}}\uparrow$}\\
       
       \midrule
       
       Tafasca~\cite{tafasca2023childplay} & \textbf{0.115} & - & - \\
       Gupta~\cite{gupta2022modular} & 0.142 & - & - \\

       \midrule
       
       MTGS~\cite{gupta2023_gazeinteract} & 0.122 & 0.588 & 0.376 \\
       \rowcolor{BackgroundColor} MTGS + AVA + CP & 0.129 & 0.586 & 0.371\\
       \rowcolor{BackgroundColor} MTGS + HICO & 0.119 & \textbf{0.601} & \underline{0.407}\\
       \rowcolor{BackgroundColor} MTGS + SWIG & \underline{0.117} & \underline{0.600} & \textbf{0.426}\\

       \bottomrule
    \end{tabular}
    }
    \caption{Cross-dataset results for the models trained on GazeFollow and evaluated on the ChildPlay dataset. Best results are given in bold, second best results are underlined.}
    \label{tab:results-childplay}
\end{table}

\begin{table}[t]
\resizebox{\linewidth}{!}{
\centering
\begin{tabular}{l | cccc}
\toprule
\textbf{Method} & \textbf{AUC$\uparrow$} & \textbf{Avg. Dist$\downarrow$}  &  \textbf{Min. Dist$\downarrow$} & \textbf{F1$_{\text{LAH}}$$\uparrow$}\\
\midrule
Multi Fusion & 0.932 & 0.119 & 0.062 & 0.633\\
Early Fusion & 0.936 & \textbf{0.118} & \textbf{0.061} & \textbf{0.643}\\
\bottomrule
\end{tabular}
}
\caption{Ablation on early vs multi-stage fusion of VLM context using the AVA+ChildPlay classes on the GazeFollow dataset. Best results are given in bold.}
\label{tab:ablation}
\vspace{-2mm}
\end{table}

\mypartitle{GazeFollow.}
We provide results for incorporating VLM context on the GazeFollow dataset in Table~\ref{tab:results-gf}. We observe that performance does not change much for the distance score. In contrast, for LAH, we observe a slight improvement with the addition of AVA+CP cues, and a degradation with the addition of SWIG cues. However, the GazeFollow test set is very small (approx. 5k instances), and often contains simple scenes with a single salient target such as the held object. Also, annotations on GazeFollow are not always reliable as mentioned in Section~\ref{sec:experiments-gf}. Hence, analyzing results on GazeFollow alone is not sufficient.

\mypartitle{ChildPlay.}
To further investigate the properties of our models, we perform cross-dataset evaluation on ChildPlay. The ChildPlay test set has a large number of instances (approx. 20k), and contains challenging scenes with multiple salient targets (ex. toys, other children/adults), making it an interesting benchmark. We observe that incorporating the AVA+CP classes results in a drop in performance for the distance score. However, with the larger set of HICO and SWIG classes, there is an improvement in performance for distance, LAH and LAEO. In particular, incorporating the SWIG classes gives the most improvements, with gaze following results comparable to the state of the art~\cite{tafasca2023childplay} and contrasts with our observations on GazeFollow. This suggests that incorporating gaze contextual cues can result in more robust performance with better generalization.

\mypartitle{Ablation: Early Fusion vs Multi-Stage Fusion.}
We perform an ablation with two different fusion mechanisms for incorporating VLM contextual information in MTGS. The first is early fusion, and follows the approach described in Section~\ref{sec:text_improved_gaze_method}. The second is a multi-stage fusion approach, where the VLM context is fused with the person tokens at every block of the architecture (4 times). We observe that the early fusion approach slightly outperforms the multi-stage fusion approach, especially for LAH, so we followed the early fusion approach for all our experiments.

\begin{figure*}[t]
    \centering
    \includegraphics[width=0.91\linewidth]{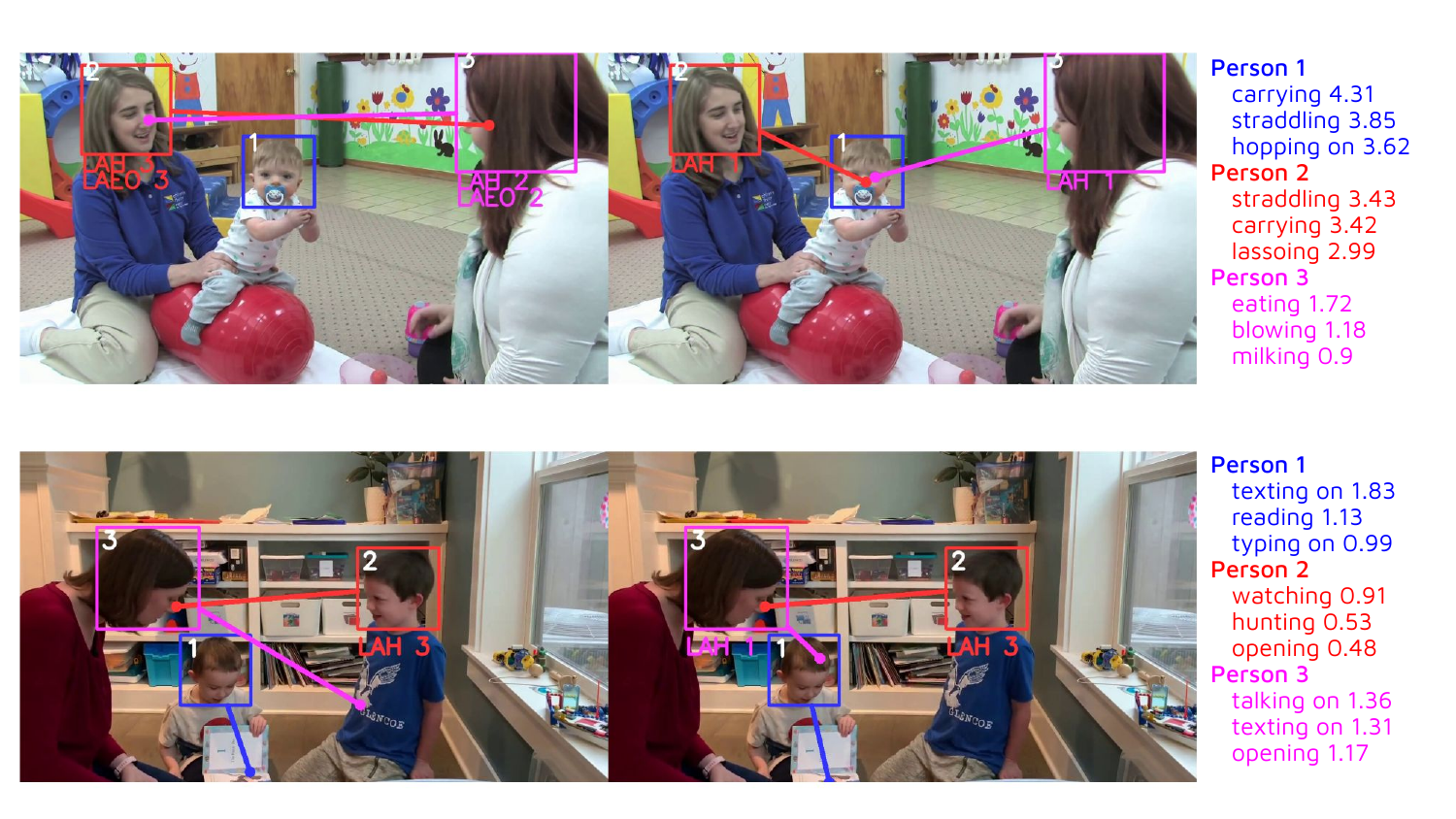}
    \caption{Qualitative results of MTGS~\cite{gupta2023_gazeinteract} trained on GazeFollow and evaluated on ChildPlay. For each person, we display the predicted gaze point as well as social gaze task along with the associated person id. We provide results without contextual cues (left) and with contextual cues from the HICO classes (right). We also display the top three classes with the highest normalized score for each person.  }
    \label{fig:example_gaze}
    \vspace{-2mm}
\end{figure*}

\mypartitle{Qualitative results.}
We provide qualitative results for MTGS, with and without the use of contextual cues in Figure~\ref{fig:example_gaze}. We observe that incorporating contextual cues can improve performance, helping identify the gaze target in challenging situations with multiple salient people and objects. For instance, in row 1, person 2 has a high score for \textit{carrying}, which might indicate that this person is looking towards their hand. In row 2, person 3 has a high score for \textit{talking on}, which suggests social interaction such as LAH.

\section{Discussion}

Our observations in Section~\ref{sec:gf-results} suggest that incorporating a larger set of contextual cues can improve generalization performance for gaze following. As the set of cues becomes larger, it can capture more specific situations (ex. unlocking, sewing in SWIG) which are usually associated with certain gaze targets. It is worth noting that increasing the number of classes has a negligible impact on computation time. As mentioned in Section~\ref{sec:contextual_cues_method}, the ITM approach processes the text prompts and images independently to obtain text and image embeddings. The final score is then a dot product of the two. Hence, all the text embeddings can be computed and saved at the start, and then used with any new image.

We also note that the set of HICO and SWIG classes utilized in our study are obtained from HOI datasets, hence, scores for the different cues could alternatively be obtained from HOI models. This is another interesting direction of investigation, but its main drawback is that the set of cues that can be considered is fixed depending on the chosen model. On the other hand, leveraging VLMs in a zero-shot manner allows us to consider any set of cues, including larger sets than the ones we considered (with a negligible impact on computation time), or more domain specific cues tailored for specific applications.

\section{Conclusion}

In this work, we explored the zero-shot capabilities of VLMs for extracting contextual cues related to a person's pose or interactions with objects and other people, and evaluated the impact of incorporating these cues into a gaze following model. We learned that VLMs can indeed extract contextual cues, and that considering the entire image with a red-circle drawn over the person of interest serves as the best visual prompt, and that ensembling scores from different textual prompts serves as the best text prompting strategy. We also observed that BLIP-2 is the overall best performing VLM, and that ICL can potentially bring further benefits. In the second part, we observed that incorporating the extracted cues into a gaze following model can provide better generalization performance, especially when considering a larger set of classes. In future work, we plan to investigate other VLMs and further explore prompting strategies such as ICL. We also plan to explore the option of predicting the different cues rather than providing them as input to the model.

\mypartitle{Acknowledgement.}
This research was supported by the AI4Autism project (digital phenotyping of autism spectrum disorders in children, grant agreement number CRSII5\_202235 / 1) of the Sinergia interdisciplinary program of the SNSF. It was also supported by \href{https://www.innosuisse.ch/inno/en/home.html}{Innosuisse}, the Swiss innovation agency, through the NL-CH Eureka Innovation project ePartner4ALL (a personalized and blended care solution with virtual buddy for child health, number 57272.1 IP-ICT).
{
    \small
    \bibliographystyle{ieeenat_fullname}
    \bibliography{main}
}

\clearpage
\setcounter{page}{1}
\maketitlesupplementary

\section{Appendix}

\subsection{Example of visual prompts}

\begin{figure*}[t]
    \includegraphics[width=\textwidth]{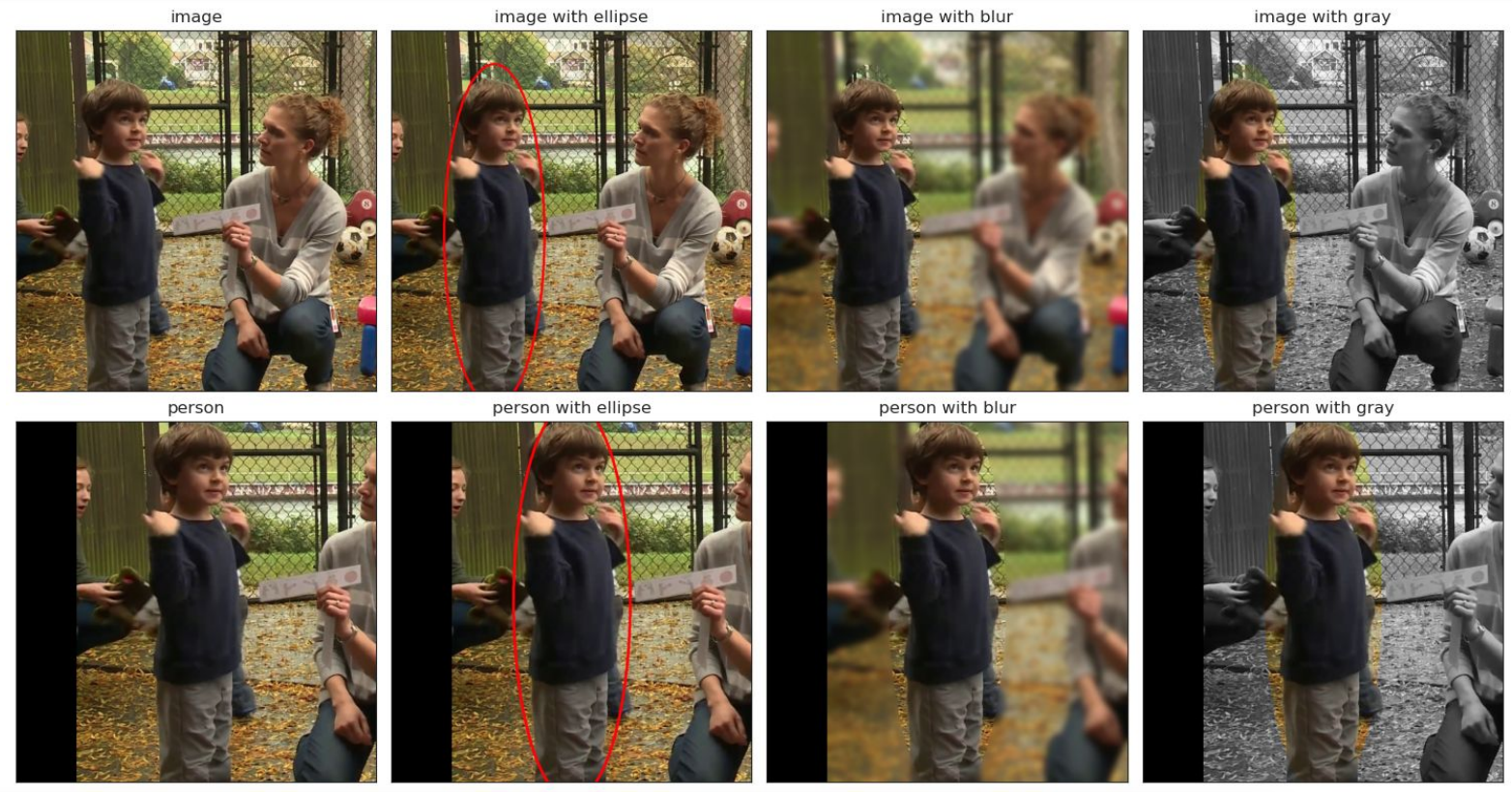}
    \caption{Different visual prompts are used to focus on the person of interest. Row-wise, the image-based and person cropped-based variants are displayed. Column-wise, various visual prompts such as ellipse, blur, and gray are presented.}
    \label{fig:visual_prompt}
\end{figure*}

As described in Section~\ref{sec:contextual_cues_method}, we investigate different visual prompting approaches to focus on a specific individual in the scene. An example of each prompt is provided in Fig.~\ref{fig:visual_prompt}. These techniques are implemented on either the whole image or specifically on the cropped image of the target person. In total, this leads to eight distinct visual prompting strategies.

\subsection{Details of the Childplay dataset}

In Table~\ref{tab:childpay_stat}, we detail the number of annotated negative and positive samples for each class in the ChildPlay dataset.

\subsection{Details of Text Prompts}

\begin{figure}[t]
    \includegraphics[width=1.0\linewidth]{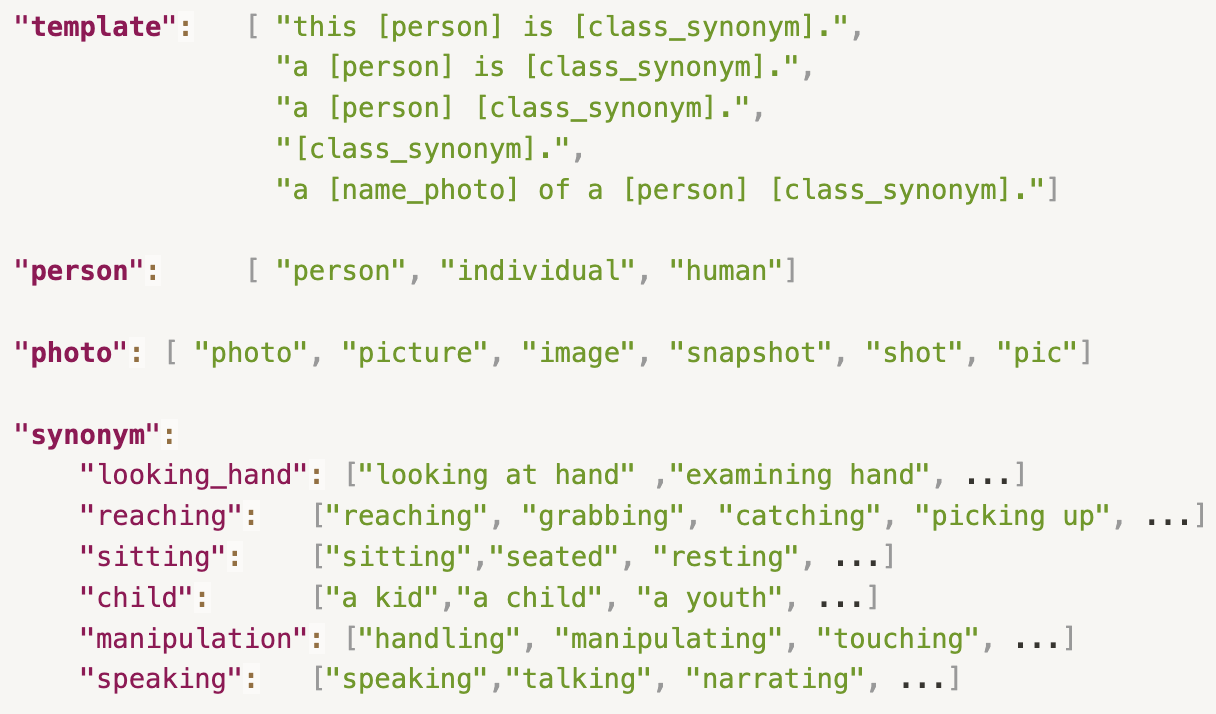}
    \caption{List of the different prompts variations used as described in section \ref{sec:text prompting}. A final prompt is a combination of \{template\},\{person\}, \{photo\} and \{synonym\} such as \textit{"this individual is grabbing"} or \textit{"a snapshot of a human handling"}. }
    \label{fig:text_prompt}
\end{figure}

\mypartitle{ITM.}
Fig.~\ref{fig:text_prompt}, lists different text prompt variations as described in Section \ref{sec:text prompting} for the ITM approach. A final prompt is a combination of \{template\},\{person\}, \{photo\} and \{synonym\} such as \textit{"this individual is grabbing"} or \textit{"a snapshot of a human handling"}.

\begin{figure}[]
    \includegraphics[width=1.0\linewidth]{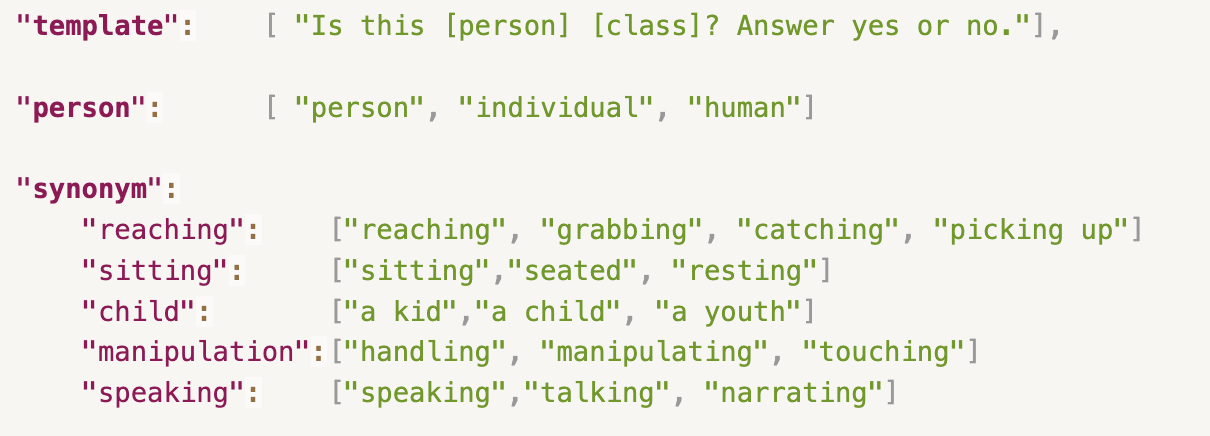}
    \caption{List of the different prompt variations used for VQA model. A final prompt is a combination of \{template\},\{person\}, and \{synonym\} such as \textit{"Is this individual grabbing? Answer yes or no."} . }
    \label{fig:vqa_prompt}
\end{figure}

\mypartitle{VQA.}
For the VQA approach, for computational reasons, we consider a single template in the form of a question, and reduce the number of synonyms for the classes. We provide the template and synonyms in Fig.~\ref{fig:vqa_prompt}.



\subsection{Impact of class synonyms}

\begin{figure*}[t]
    \includegraphics[width=1.0\linewidth]{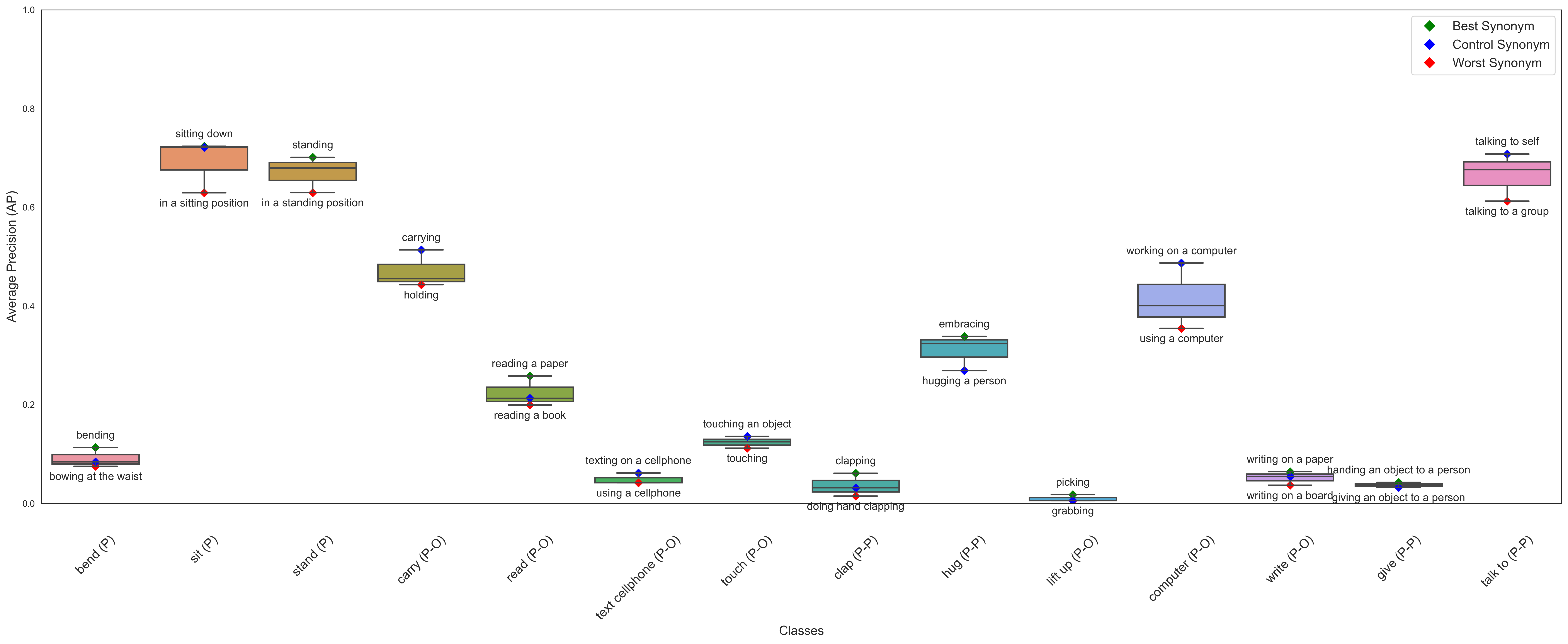}
    \caption{Performance when varying the class synonym in the text prompt. We display the mean and variance of results, as well as the best and worst synonym.}
    \label{fig:synonym}
\end{figure*}

In Fig.~\ref{fig:synonym}, we provide the results for varying the class synonym in the text prompt. We observe that performance can change depending on the used synonym by a large margin.

\textcolor{white}{Contrary to popular belief, Lorem Ipsum is not simply random text. It has roots in a piece of classical Latin literature from 45 BC, making it over 2000 years old. Richard McClintock, a Latin professor at Hampden-Sydney College in Virginia, looked up one of the more obscure Latin words, consectetur, from a Lorem Ipsum passage, and going through the cites of the word in classical literature, discovered the undoubtable source. Lorem Ipsum comes from sections 1.10.32 and 1.10.33 of "de Finibus Bonorum et Malorum" (The Extremes of Good and Evil) by Cicero, written in 45 BC. This book is a treatise on the theory of ethics, very popular during the Renaissance. The first line of Lorem Ipsum, "Lorem ipsum dolor sit amet..", comes from a line in section 1.10.32.}

\begin{table}[t]
\resizebox{\linewidth}{!}{
    \centering
    \begin{tabular}{L{4cm}  M{1.5cm} M{1.5cm} }
        \toprule
        \textbf{Classes} & \textbf{negative} & \textbf{positive} \\
        \hline
        looking at hand & 36 & 35 \\
        reaching & 36 & 34 \\
        sitting & 60 & 52 \\
        child & 59 & 58 \\
        manipulation & 59 & 59 \\
        speaking & 31 & 30 \\
        \bottomrule
    \end{tabular}
    }
    \caption{Classes and statistics of the ChildPlay dataset annotation.}
    \label{tab:childpay_stat}
\end{table}

\end{document}